\documentclass{article}

\usepackage{arxiv}

\usepackage[utf8]{inputenc} 
\usepackage[T1]{fontenc}    
\usepackage{hyperref}       
\usepackage{url}            
\usepackage{booktabs}       
\usepackage{amsfonts}       
\usepackage{nicefrac}       
\usepackage{microtype}      
\usepackage{lipsum}
\usepackage{cite}
\usepackage{amsmath,amssymb,amsfonts}
\usepackage{algorithmic}
\usepackage{graphicx}
\usepackage{textcomp}
\usepackage{tikz}
\usepackage{caption}
\usepackage{enumitem}
\usepackage{subfigure}
\usepackage{booktabs}
\usepackage{bm}
\usepackage{hyperref}

\title{Highly Fast Text Segmentation With Pairwise Markov Chains}

\author{
Elie Azeraf\thanks{Elie Azeraf is also a member of SAMOVAR, Telecom SudParis, Institut Polytechnique de Paris} \\
  Watson Department \\
  IBM GSB France \\
  \texttt{elie.azeraf@ibm.com} \\
   \And
Emmanuel Monfrini \\
SAMOVAR, Telecom SudParis \\
Institut Polytechnique de Paris \\
\And
Emmanuel Vignon \\
Watson Department \\
  IBM GSB France \\
\And
Wojciech Pieczynski \\
SAMOVAR, Telecom SudParis \\
Institut Polytechnique de Paris \\
}

\begin{document}
\maketitle

\begin{abstract}
Natural Language Processing (NLP) models' current trend consists of using increasingly more extra-data to build the best models as possible. It implies more expensive computational costs and training time, difficulties for deployment, and worries about these models' carbon footprint reveal a critical problem in the future. Against this trend, our goal is to develop NLP models requiring no extra-data and minimizing training time. To do so, in this paper, we explore Markov chain models, Hidden Markov Chain (HMC) and Pairwise Markov Chain (PMC), for NLP segmentation tasks. We apply these models for three classic applications: POS Tagging, Named-Entity-Recognition, and Chunking. We develop an original method to adapt these models for text segmentation's specific challenges to obtain relevant performances with very short training and execution times. PMC achieves equivalent results to those obtained by Conditional Random Fields (CRF), one of the most applied models for these tasks when no extra-data are used. Moreover, PMC has training times 30 times shorter than the CRF ones, which validates this model given our objectives.
\end{abstract}


\section{Introduction}

In the past ten years, developments of Deep Learning methods \cite{Goodfellow-et-al-2016, lecun2015deep} have enabled research in Natural Language Processing (NLP) to take an impressive leap. Some tasks, like Question Answering \cite{rajpurkar2016squad} or Sentiment Analysis \cite{maas-EtAl:2011:ACL-HLT2011}, seemed unrealistic twenty years ago, and nowadays recent neural models achieved better scores than humans \cite{devlin2018bert} \cite{lan2019albert} for these applications. This dynamic's main motivation is the direct applications of NLP models in the industry, with tasks such as Named-Entity-Recognition and mail classification. However, the cost of improving models' performances increases, and learning algorithms always need more data and power to be trained and make a prediction. We can produce models that achieve impressive scores, but deployment and climate change problems \cite{strubell2019energy} raise some issues. Our aim in this paper is to initiate a reflection around light models by introducing a new Markov model design for text segmentation and comparing it with machine learning algorithms that have a reasonable carbon impact and execution time. Motivation about the choice of segmentation tasks is explained later in this paper.

The Hidden Markov Chains (HMC), introduced by Stratonovitch sixty years ago \cite{stratonovich1965conditional} \cite{baum1966statistical} \cite{cappe2006inference} \cite{rabiner1986introduction} \cite{rabiner1989tutorial} \cite{ephraim2002hidden}, which model poor correlations, are widely used in machine learning, and have especially been applied for NLP segmentation tasks \cite{brants2000tnt} \cite{morwal2012named} \cite{ekbal2007pos}.

For over twenty years, HMCs have been strictly generalized to Pairwise Markov Chains (PMCs) \cite{pieczynski2003pairwise} \cite{gorynin2016optimal}, a family of models including HMCs. In PMCs, the ``hidden" chain is not necessarily Markov, and the noise is modeled in a more correlated – and thus more informative – way. However, PMCs keep the same advantages as HMCs in regards to the hidden data estimation. In particular, the training and estimation tasks still have linear complexity. PMCs have especially been studied for image segmentation, with discrete hidden variables and continuous observations. It turns out that using PMCs instead of HMCs can divide the error rate by two \cite{derrode2004signal} \cite{gorynin2018assessing}. However, for specific reasons relative to language processing, which we will develop later, PMCs have never been applied for NLP tasks.

This paper explores the interest in using PMCs for three of the main text segmentation tasks: Part-Of-Speech (POS) Tagging, Chunking, and Named-Entity-Recognition (NER). The best methods for these tasks are based on Deep Learning models \cite{yang2019xlnet} \cite{akbik2018coling}. However, to produce excellent scores, these models require a large amount of extra-data, which results in very long training time and difficulties in deploying them with classic architectures.

The paper is organized as follows. In the next section, we present and compare PMCs and HMCs, the bayesian segmentation methods, and the parameter estimation algorithm. The third section is devoted to the text segmentation tasks and an original way to adapt Markov chain models for these tasks while keeping fast training and relevant results. Experiments are presented in section four. The last section is devoted to conclusions and perspectives. 

\section{Markov Chain Models}

Let $X_{1:T} = (X_1, ..., X_T)$ and $Y_{1:T} = (Y_1, ..., Y_T)$ be two discrete stochastic processes. For all $t$ in $\{1, ..., T\}$, $X_t$ takes its values in $\Lambda_X = \{\lambda_1, ..., \lambda_N\}$ and $Y_t$ takes its values in $\Omega_Y = \{\omega_1, ..., \omega_M\}$. Let us then consider the process $Z_{1:T} = (X_{1:T}, Y_{1:T})$.

Our study takes place in the case of latent variables, with an observed realization $y_{1:T}$ of $Y_{1:T}$, and the corresponding hidden realization $x_{1:T}$ of $X_{1:T}$ having to be estimated.

In the following, in order to simplify the notation, for all $t$ in $\{1, ..., T\}$, events $\left\{X_t=x_t\right\}$, $\left\{Y_t=y_t\right\}$ and $\left\{Z_t=z_t\right\}$ will be denoted by $\{x_t\}$, $\{y_t\}$ and $\{z_t\}$, respectively.

\subsection{HMC - PMC}

Starting from the most correlated shape of the couple $(X_{1:T}, Y_{1:T})$, we can say that $Z_{1:T}$ is an HMC if it is possible to write its probability law as:
\begin{equation}
\begin{split}
    p(z_{1:T}) = p(x_1) p(y_1 | x_1)& p(x_2 | x_1) 
    p(y_2 | x_2) \\ &... p(x_T | x_{T - 1}) p(y_T | x_T)
    \end{split}
\label{eq:HMC}
\end{equation}

Moreover, for all $\lambda_i, \lambda_j \in \Lambda_X$ and $\omega_k \in \Omega_Y$, we consider homogeneous HMCs defined with the following parameters:
\begin{itemize}
    \item Initial probability $\pi = (\pi(1), ..., \pi(N))$ with \\
    $\pi(i) = p(x_1 = \lambda_i)$
    \item Transition matrix $A = \{ a_{i}(j) \}$ with \\ 
    $\forall t \in \{ 1, ..., T - 1 \}, a_{i}(j) = p(x_{t+1} = \lambda_j | x_{t} = \lambda_i)$ 
    \item Emission matrix $B = \{ b_{i}(k) \}$ with \\
    $\forall t \in \{ 1, ..., T \}, b_{i}(k) = p(y_t = \omega_k | x_t = \lambda_i)$
\end{itemize}

\begin{figure}[h]
\begin{center}
\begin{tikzpicture}
[font=\small, inner sep=0pt, hidden/.style = {circle,draw = blue!15, fill = blue!10, thick,minimum size = 0.75cm, rounded corners}, visible/.style = {circle,draw=black!35,fill=black!30,thick,minimum size=0.75cm, rounded corners}, scale = 0.75]
\node at (-5,0) (x1) [visible] {$x_1$};
\node at (-5,2) (h1) [hidden]  {$y_1$};
\draw [->, >=stealth] (x1) to (h1);
                    
\node at (-3,0) (x2) [visible] {$x_2$};
\node at (-3,2) (h2) [hidden]  {$y_2$};
\draw [->, >=stealth] (x2) to (h2);
                    
\node at (-1,0) (x3) [visible] {$x_3$};
\node at (-1,2) (h3) [hidden]  {$y_3$};
\draw [->, >=stealth] (x3) to (h3);
                    
\node at (1,0) (x4) [visible] {$x_4$};
\node at (1,2) (h4) [hidden]  {$y_4$};
\draw [->, >=stealth] (x4) to (h4);
                    
\draw [->, >=stealth] (x1) to (x2);
\draw [->, >=stealth] (x2) to (x3);
\draw [->, >=stealth] (x3) to (x4);
\end{tikzpicture}
\end{center}
\captionof{figure}{Probabilistic graphical model of HMC}
\label{fig:HMC}
\end{figure}
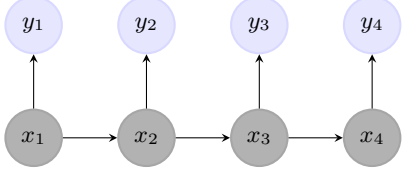

An oriented dependency graph of the HMC is given in figure \ref{fig:HMC}.

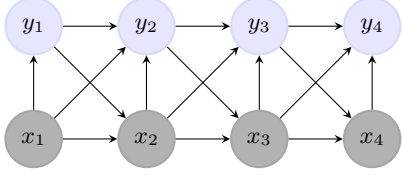
\begin{figure}[h!]
\begin{center}
\begin{tikzpicture}
[font=\small, inner sep=0pt, hidden/.style = {circle,draw = blue!15, fill = blue!10, thick,minimum size = 0.75cm, rounded corners}, visible/.style = {circle,draw=black!35,fill=black!30,thick,minimum size=0.75cm, rounded corners}, scale = 0.75]
\node at (-5,0) (x1) [visible] {$x_1$};
\node at (-5,2) (y1) [hidden]  {$y_1$};
\draw [->, >=stealth] (x1) to (y1);
                    
\node at (-3,0) (x2) [visible] {$x_2$};
\node at (-3,2) (y2) [hidden]  {$y_2$};
\draw [->, >=stealth] (x2) to (y2);
                    
\node at (-1,0) (x3) [visible] {$x_3$};
\node at (-1,2) (y3) [hidden]  {$y_3$};
\draw [->, >=stealth] (x3) to (y3);
                    
\node at (1,0) (x4) [visible] {$x_4$};
\node at (1,2) (y4) [hidden]  {$y_4$};
\draw [->, >=stealth] (x4) to (y4);
                    
\draw [->, >=stealth] (x1) to (x2);
\draw [->, >=stealth] (x2) to (x3);
\draw [->, >=stealth] (x3) to (x4);

\draw [->, >=stealth] (x1) to (y2);
\draw [->, >=stealth] (x2) to (y3);
\draw [->, >=stealth] (x3) to (y4);

\draw [->, >=stealth] (y1) to (x2);
\draw [->, >=stealth] (y2) to (x3);
\draw [->, >=stealth] (y3) to (x4);

\draw [->, >=stealth] (y1) to (y2);
\draw [->, >=stealth] (y2) to (y3);
\draw [->, >=stealth] (y3) to (y4);

\end{tikzpicture}
\end{center}
\captionof{figure}{Probabilistic graphical model of PMC}
\label{fig:PMC}
\end{figure}

We can note that the dependency graph is almost reduced to the minimum to link dependent couple of sequential data, and yet HMC is known to be a robust model. PMCs, which we will detail just after, allow to introduce more connections to the graph, as shown in figure \ref{fig:PMC}.

Starting from the most correlated shape of the couple $(x_{1:T}, y_{1:T})$, we can say that $z_{1:T}$ is a PMC if $z_{1:T}$ is a Markov chain, which is equivalent to admitting that its distribution is of the form:
\begin{equation}
p(z_{1:T}) = p(z_1) p(z_2 | z_1)...p(z_T | z_{T - 1})
\end{equation}
which can be rewritten, without loss of generality,
\begin{equation}
\begin{split}
\hspace{-1cm} p(z_{1:T}) =& p(x_1) p(y_1 |x_1) p(x_2| x_1, y_1) p(y_2| x_2, x_1, y_1) \\ &...p(x_T| x_{T - 1}, y_{T - 1}) p( y_T |x_T, x_{T - 1}, y_{T - 1})
\end{split}
\label{eq:PMC}
\end{equation}

When comparing (\ref{eq:HMC}) and (\ref{eq:PMC}) we can see that a PMC is a HMC \cite{pieczynski2003pairwise} if and only if $\forall t \in \{1,...,T-1\}$,
\begin{itemize}
    \item $p(x_{t+1}| x_t, y_t) = p(x_{t+1}| x_t)$ and
    \item $p(y_{t+1}| x_{t+1}, x_t, y_t) = p(y_{t+1}| x_{t+1})$.
\end{itemize}

From a graphical point of view, we can materialize on figures \ref{fig:HMC} and \ref{fig:PMC} how PMCs are more general than HMCs. In particular, the markovian assumption of $x_{1:T}$ made for HMCs is not supposed for PMCs, but the determinist bayesian inference is still possible.

In this paper, we consider homogeneous PMCs defined with three sets of parameters, $\forall \lambda_i, \lambda_j \in \Lambda_X, \forall \omega_k, \omega_l \in \Omega_Y, \forall t \in \{1, ..., T - 1 \}$:
\begin{itemize}
    \item Initial probability matrix $\Pi^{PMC} = \{ \pi^{PMC}(i, k) \}$ with \\
    $\pi^{PMC}(i, k) = p(x_1 = \lambda_i, y_1 = \omega_k)$;
    \item Transition matrix 
    $A^{PMC} = \{ a^{PMC}_{i,k}(j) \}$ with: \\
    $a^{PMC}_{i, k}(j) = p(x_{t+1} = \lambda_j | x_{t} = \lambda_i, y_{t} = \omega_k) $;
    \item Emission matrix $B^{PMC} = \{ b^{PMC}_{i, j, k}(l) \}$: \\ 
    $b^{PMC}_{i, j, k}(l) = p(y_{t+1} = \omega_l | x_{t} = \lambda_i, x_{t+1} = \lambda_j, y_{t} = \omega_k)$.
\end{itemize}

\subsection{Bayesian segmentation}

There are two Bayesian methods to estimate the realization $\hat{x}_{1:T}$ of the hidden chain: the Marginales Posterior Mode (MPM) and the Maximum A Posteriori (MAP).

The MPM estimator is given by :
\begin{equation*}
\begin{split}
& \hat{x}_{MPM} = (\hat{x}_1, ..., \hat{x}_T) \text{ with } \forall t\in\{1, ..., T\},\\ 
& p(\hat{x}_t | y_{1:T}) = \sup_{\lambda_i \in \Lambda_X} p(x_t = \lambda_i | y_{1:T})
\end{split}
\end{equation*}
and the MAP estimator is :
\begin{equation*}
\hat{x}_{MAP} = \text{arg}\max_{x_{1:T} \in (\Lambda_X)^T} p(x_{1:T} | y_{1:T})
\end{equation*}
MPM estimator is given with Forward-Backward algorithm \cite{baum1966statistical} \cite{rabiner1989tutorial}, while MAP is given by the Viterbi one \cite{viterbi1967error}. We tested both of them, and MPM gives slightly better results than MAP in term of accuracy for the tasks we consider, so we only present results for MPM.

We thus have to compute $\forall t \in \{1, ..., T \}, \forall \lambda_i \in \Lambda_X$, $p(x_t = \lambda_i | y_{1:T})$ to maximize posterior marginals with the most probable state at every time $t \in \{1,...,T\}$. We present the Forward-Backward algorithm in the case of PMC which generalizes the classical HMC one.

$\forall t \in \{1, ..., T \}, \forall \lambda_i \in \Lambda_X$:
\begin{equation*}
    p(x_t = \lambda_i | y_{1:T}) = \frac{\alpha_t(i) \beta_t(i)}{\sum_{j \in \Lambda_X} \alpha_t(j) \beta_t(j)}
\end{equation*}
where $\forall t\in\{1,...,T\}, \forall \lambda_i \in \Lambda_X$:
\begin{equation*}
\begin{split}
&\alpha_t(i) = p(y_{1:t}, x_t = \lambda_i) \\
&\beta_t(i) = p(y_{t + 1:T}| x_t = \lambda_i, y_t)
\end{split}
\end{equation*}

$\forall \lambda_i \in \Lambda_X, \forall t \in \{ 1, ..., T \}, \alpha_t(i)$ and $\beta_t(i)$ can still be computed thanks to forward and backward recursions:
\begin{itemize}
    \item with $y_1 = \omega_k$, 
    \begin{align*}
    \alpha_1(i) = \pi^{PMC}(i, k) 
    \end{align*}
    \item $\forall t\in\{1,..., T - 1 \}$, with $y_{t} = \omega_k, y_{t+1} = \omega_l$,
    \begin{align*}
        \alpha_{t+1}(i) = \sum_{\lambda_j \in \Lambda_X} a^{PMC}_{j,i}(k) b^{PMC}_{j, i, k}(l)  \alpha_{t}(j)
    \end{align*}
\end{itemize}
and
\begin{itemize}
    \item $\beta_T(i) = 1$
    \item $\forall t \in \{1,...,T - 1\}$  with  $y_{t} = \omega_k, y_{t+1} = \omega_l$,
    \begin{align*}
    \beta_t(i) = \sum_{\lambda_j \in \Lambda_X} a^{PMC}_{i,j}(k) b^{PMC}_{i, j, k}(l) \beta_{t + 1}(j)
    \end{align*}
\end{itemize}

One can normalize forward and backward probabilities at every step, which allows them to avoid underflowed computational problems without modifying the results. This algorithm can be executed with matrix computation, allowing a highly fast execution.

\subsection{Parameter estimation}

We estimate the parameters with maximum likelihood estimator \cite{Jurafsky:2000:SLP:555733} \cite{brants2000tnt}. It consists in computing the empirical frequencies of the probabilities of interest. We have, $\forall \lambda_i, \lambda_j \in \Lambda_X,\forall \omega_k, \omega_l \in \Omega_Y,$
\begin{align*}
&\hat{\pi}(i) = \frac{N^0_{i}}{L},~~\hat{a}_{i}(j) = \frac{N_{i,j}}{N_i},~~\hat{b}_{i}(k) = \frac{M_{i,k}}{N_i}\\
\text{and  }~~&\hat{\pi}^{PMC}(i, k) = \frac{N^0_{i,k}}{L},~~\hat{a}_{i,k}^{PMC}(j) = \frac{N_{i,k,j}}{M_{i,k}},\\
&\hat{b}_{i,j,k}^{PMC}(l) = \frac{N_{i,k,j,l}}{N_{i,k,j}}
\end{align*}

\noindent where $N_{i,k,j,l}$ is the number of occurrences of the pattern $(x_t = \lambda_i, y_t = \omega_k, x_{t+1} = \lambda_j, y_{t+1} = \omega_l)$ in the $L$ chains of the training set.
Then,  $N_{i,k,j} = \sum_{\omega_l \in \Omega_Y}N_{i,k,j,l}$,   $N_{i,j} = \sum_{\omega_k \in \Omega_Y} N_{i,k,j}$, $M_{i,k} = \sum_{\lambda_j \in \Lambda_X} N_{i,k,j}$ and $N_{i} = \sum_{\lambda_j \in \Lambda_X} N_{i,j}$. Finally, $N^0_i$ and $N^0_{i,k}$ are respectively the number of times $x_1=\lambda_i$ and $z_1=(\lambda_i, \omega_k)$ in the $L$ chains of the training set.

\begin{figure}[h]
\begin{center}
\begin{tikzpicture}
[font=\small, inner sep=0pt, hidden/.style = {circle,draw = blue!15, fill = blue!10, thick,minimum size = 0.75cm, rounded corners}, visible/.style = {circle,draw=black!35,fill=black!30,thick,minimum size=0.75cm, rounded corners}, scale = 0.75]
\node at (-5,0) (x1) [visible] {$x_1$};
\node at (-5,2) (y1) [hidden]  {$y_1$};
\draw [->, >=stealth] (x1) to (y1);
                    
\node at (-3,0) (x2) [visible] {$x_2$};
\node at (-3,2) (y2) [hidden]  {$y_2$};
\draw [->, >=stealth] (x2) to (y2);
                    
\node at (-1,0) (x3) [visible] {$x_3$};
\node at (-1,2) (y3) [hidden]  {$y_3$};
\draw [->, >=stealth] (x3) to (y3);
                    
\node at (1,0) (x4) [visible] {$x_4$};
\node at (1,2) (y4) [hidden]  {$y_4$};
\draw [->, >=stealth] (x4) to (y4);

\node at (3,0) (x5) [visible] {$x_5$};
\node at (3,2) (y5) [hidden]  {$y_5$};
\draw [->, >=stealth] (x5) to (y5);

\node at (5,0) (x6) [visible] {$x_6$};
\node at (5,2) (y6) [hidden]  {$y_6$};
\draw [->, >=stealth] (x6) to (y6);
                    
\draw [->, >=stealth] (x1) to (x2);
\draw [->, >=stealth] (x2) to (x3);
\draw [->, >=stealth] (x3) to (x4);
\draw [->, >=stealth] (x4) to (x5);
\draw [->, >=stealth] (x5) to (x6);

\draw [->, >=stealth] (x1) to (y2);
\draw [->, >=stealth] (x3) to (y4);
\draw [->, >=stealth] (x4) to (y5);

\draw [->, >=stealth] (y1) to (x2);
\draw [->, >=stealth] (y3) to (x4);
\draw [->, >=stealth] (y4) to (x5);

\draw [->, >=stealth] (y1) to (y2);
\draw [->, >=stealth] (y3) to (y4);
\draw [->, >=stealth] (y4) to (y5);

\end{tikzpicture}
\captionof{figure}{Graphical model of a partially ``downgraded'' PMC - The couple of observations $(y_2, y_3)$ and $(y_5, y_6)$ never appeared in this order in the training set}
\label{fig:MPCHMC}
\end{center}
\end{figure}
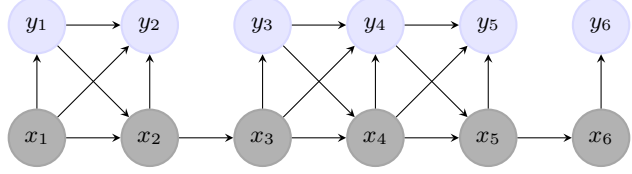

When $\text{card}(\Lambda_X)$ or $\text{card}(\Omega_Y)$ are huge, some patterns may not be observed in the training set, which implies that the corresponding estimation is zero. It is the case for NLP with $\Omega_Y$, the space of possible written words. To minimize this problem, especially in PMC, we accept to partially ``downgrade" PMC to HMC when necessary. This original process is represented in figure \ref{fig:MPCHMC}. It consists of approximating the forward and backward probabilities of the PMC by the HMC ones when those of PMC equal $0$.

Moreover, it is essential to note that ``online'' learning is fast and easy in those models. If new sentences enrich the training set, updating parameters only requires us to ``add'' information to already determined values without retraining the model on the complete dataset.

\section{Text segmentation tasks with Markov Chain models}

Let us now have a few words on the three text segmentation tasks that we are interested in:  POS Tagging, Chunking, and NER. For those three tasks, we observe the words of some sentences for which we have to find the specific labels.

POS tagging consists in labeling each word with its grammatical function as verb, determinant, adjective... For example, $(y_1, y_2, ..., y_{12}) = $ (\textit{John, likes, the, blue, house, at, the, end, of, the, street, .}) and $(x_1, x_2, ..., x_{12}) =  $ (\textit{Noun, Verb, Det, Adj, Noun, Prep, Det, Noun, Prep, Det, Noun, Punct}). The performance of POS Tagging is evaluated with the accuracy error.

NER consists in discriminating the ``entities'' from the words of sentences. The ``entities'' can be a person (PER), a localization (LOC), or an organization (ORG). The set of entities depends on the use case. For example, (\textit{John, works, at, IBM, in, Paris, .}) can have the following entities (\textit{PER, O, O, ORG, O, LOC, O}) where \textit{O} stands for ``no entity''. The set of entities depends on the use case, for example, finding the name of protein or DNA in medical data as in \cite{ohta2002genia}. The performance for NER is evaluated with $F_1$ score \cite{derczynski2016complementarity}. 

Chunking consists of segmenting a sentence into its sub-constituents, such as noun (NP), verb (VP), and prepositional phrases (PP). For example, (\textit{We, saw, the, yellow, dog, .}) has the chunks: (\textit{NP, VP, NP, NP, NP, O}). Like NER, the performance is evaluated with the $F_1$ score. More details about these tasks can be found in the NLTK book \cite{loper2002nltk}.

We took these three tasks as they can be the basis of more complex ones. For example, with text classification when the corpus is small, with limited architecture avoiding the deployment of a heavy deep learning model. This case can happen, for example, with email classification, with models having to process hundreds or thousands of emails per second with a maximum RAM power of 4Gb, given a relatively small training set. One way to handle this problem is to construct a text segmentation model, then another one using the predicted labels, to make a prediction. The three label types we consider, POS tag, chunk tag, and entity, are particularly useful to construct this type of architecture, motivating the choice of these tasks.

However, When working in the context of text segmentation, the cardinal of $\Omega_Y$ is very large, and managing unknown patterns (not in the train set) is then a real challenge. Although the Markov properties of our processes partially help to tackle this problem, it is not enough. PMC and HMC have to be adapted to consider this problem. Our goal is also to keep a very short training time. First of all, we can use the ``downgrading" process from PMC to HMC described in Section II.B to alleviate this problem when PMC is used.  

Moreover, for HMC, we look for extra information in the unknown observed words. Given a word $\omega$, We introduce the following functions:
\begin{itemize}
\item $u(\omega) = 1$ if the first letter of $\omega$ is up, $0$ otherwise
\item $h(\omega) = 1$ if $\omega$ has a hyphen, $0$ otherwise
\item $f(\omega) = 1$ if $\omega$ is the first word of the sentence, $0$ otherwise
\item $d(\omega) = 1$ if $\omega$ has a digit, 0 otherwise
\item $s_m(\omega) = \text{the suffix of length m of } \omega$
\end{itemize}
Then, for $\omega_k\in \Omega_Y$ unknown, and $\forall \lambda_i \in \Lambda_X$, we approximate $b_{i,k}$ by
$$ b_{i}(k) \approx p(u(\omega_k), h(\omega_k), f(\omega_k), d(\omega_k), s_3(\omega_k) |x_t = \lambda_i)$$ 
which is also estimated with empirical frequencies. If $s_3(\omega_k)$ is unknown in the training set, we use $s_2(\omega_k)$ or otherwise $s_1(\omega_k)$ or finally $s_0(\omega_k)$. This original approximation method allows us to improve the segmentation of unknown words while keeping a very short training time with PMC and HMC. The choice of the different features depends of the language, these ones are selected for English, and one can add any features according to language characteristics. 

It is important to note that the choice of the features is crucial as we cannot use arbitrary features with Markov chain models \cite{Jurafsky:2000:SLP:555733} \cite{mccallum2000maximum} \cite{sutton2006introduction}. It preserves the model from a ``second level'' of unknown patterns, and it avoids having our new approximation of $b_{i}(k)$ to be equal to $0$ too many times.

\section{Experiments}

To calibrate the performance of our models, we compare their performances to benchmark models with no extra-data: Maximum Entropy Markov Models \cite{mccallum2000maximum}, Recurrent Neural Networks \cite{jozefowicz2015empirical} \cite{lipton2015critical}, Long-Short-Term-Memory (LSTM) network \cite{hochreiter1997long}, Gated Recurrent Unit \cite{chung2014empirical}, Conditional Random Field (CRF) \cite{lafferty2001conditional}, and the BiLSTM-CRF \cite{huang2015bidirectional}. We present the results of the best one with appreciable training and execution time, the CRF, having the same features described in section 3, to which we add the suffix of length 4 and prefixes of length 1 to 4. Comparing results to this model is particularly relevant, as it is one of the most applied models for these tasks when no extra-data are used. The experiments are done with python. We code our own Markov Chain models, and we use the CRF Suite \cite{CRFsuite} library for the CRF models.

\begin{table}[h]
\begin{center}
\caption{HMC, CRF and PMC for POS Tagging with error rates, KW stands for Known Words, and UW for Unknown Words}
\begin{sc}
\begin{tabular}{lccc}
\toprule
{} & HMC & CRF & PMC \\
\midrule
CoNLL 2000 & $2.96\%$ & $2.47\%$ & $\bm{2.32\%}$ \\
CoNLL 2000 KW & $1.94\%$ & $1.72\%$ & $\bm{1.27\%}$ \\
CoNLL 2000 UW & $16.54\%$ & $\bm{12.47\%}$ & $16.41\%$ \\
\midrule
CoNLL 2003 & $5.29\%$ & $\bm{4.28\%}$ & $4.71\%$ \\
CoNLL 2003 KW & $4.03\%$ & $\bm{3.21\%}$ & $3.40\%$ \\
CoNLL 2003 UW & $15.30\%$ & $\bm{12.79\%}$ & $15.16\%$ \\
\midrule
UD English & $8.13\%$ & $\bm{6.95\%}$ & $7.16\%$ \\
UD English KW & $6.07\%$ & $5.46\%$ & $\bm{5.00\%}$ \\
UD English UW & $30.73\%$ & $\bm{23.25\%}$ & $30.78\%$ \\
\bottomrule
\end{tabular}
\end{sc}
\end{center}
\vskip 0.1in
\end{table}

\begin{table}[h]
\vskip -0.1in
\begin{center}
\caption{HMC, CRF and PMC for NER with $F_1$ scores, KW stands for Known Words, and UW for Unknown Words}
\begin{sc}
\begin{tabular}{lccc}
\toprule
{} & HMC & CRF & PMC \\
\midrule
CoNLL 2003 & $78.44$ & $79.16$ & $\bm{79.52}$ \\
CoNLL 2003 KW & $87.07$ & $87.41$ & $\bm{88.47}$ \\
CoNLL 2003 UW & $56.67$ & $\bm{58.67}$ & $56.87$ \\
\bottomrule
\end{tabular}
\end{sc}
\end{center}
\vskip 0.1in
\end{table}

\begin{table}[h!]
\vskip -0.1in
\begin{center}
\caption{HMC, CRF and PMC for Chunking with $F_1$ scores, KW stands for Known Words, and UW for Unknown Words}
\begin{sc}
\begin{tabular}{lccc}
\toprule
{} & HMC & CRF & PMC \\
\midrule
CoNLL 2000 & $92.72$ &  $93.05$ & $\bm{94.49}$ \\
CoNLL 2000 KW & $93.18$ & $93.35$ & $\bm{95.09}$ \\
CoNLL 2000 UW & $87.45$ & $\bm{89.63}$ & $87.58$\\
\midrule
CoNLL 2003 & $94.30$ & $94.79$ & $\bm{95.61}$ \\
CoNLL 2003 KW & $94.65$ & $94.92$ & $\bm{96.17}$ \\
CoNLL 2003 UW & $91.95$ & $\bm{93.88}$ & $91.85$ \\
\bottomrule
\end{tabular}
\end{sc}
\end{center}
\vskip 0.1in
\end{table}

\begin{table}[h!]
\vskip -0.1in
\begin{center}
\caption{Training times of the different models on CPU with 8 Gb of RAM}
\begin{sc}
\begin{tabular}{lccc}
\toprule
{} & HMC & CRF & PMC \\
\midrule
CoNLL 2000 POS & 1.3s & 140s & 5s \\
CoNLL 2000 Chunk & 1.3s & 140s & 5s \\
\midrule
CoNLL 2003 POS & 1.5S & 180s & 5.5s \\
CoNLL 2003 NER  & 1.5s & 260s & 5.5s \\
CoNLL 2003 Chunk & 1.5s & 160s & 5.5s \\
\midrule
UD English POS & 1.7s & 185s & 6.3s \\
\bottomrule
\end{tabular}
\end{sc}
\end{center}
\vskip 0.1in
\end{table}

We use reference datasets for every tasks: CoNLL 2000 \cite{sang2000introduction} for Chunking, UD English \cite{nivre2016universal} for POS Tagging, and CoNLL 2003 
\cite{sang2003introduction} for NER \footnote{All these datasets are freely available. CoNLL 2003 upon request at \href{https:/www.clips.uantwerpen.be/conll2003//ner/}{https:/www.clips.uantwerpen.be/conll2003/ner/}, UD English is available on \href{https:/universaldependencies.org/\#language-}{https:/universaldependencies.org/\#language-}, and CoNLL 2000 with the NLTK \cite{loper2002nltk} library with python.}. In addition, we take advantage of having enough data to use CoNLL 2000 and CoNLL 2003 for POS Tagging, and CoNLL 2003 for chunking. We use universal tagset \cite{petrov2011universal} for each experiment of POS Tagging. 

The results are presented in Table I for POS Tagging, Table II for NER, and Table III for Chunking. Training times on CPU with 8 Gb of RAM are presented in Table IV. HMC has the lowest training times, but its performances are significantly worse than those of the other models. The difference can achieve 20\%. As HMC is a particular case of PMC, these results were expected. They illustrate the interest in using the PMC model rather than the HMC one. The most relevant results are those regarding the comparison between PMC and CRF. They have equivalent scores, with PMC having better results in 4 of the 6 experiments. In general, PMC is better for known words and has worse results for unknown ones. The main difference between the two models rests on the time required to train them. PMC is about 30 times faster to train than CRF, and adding new data does not entail the retraining of models. About execution time, we have the same observation, with PMC about ten times faster. It is our work's central goal: to have the best and fastest segmentation model as possible with no extra-data, and PMC seems to achieve this objective. 

\section{Conclusion}

PMC has good performances with no extra-data for these tasks, with equivalent results to the CRF. The main advantage of PMC is its training and execution time, much faster for text segmentation. It confirms the interest of using PMC to build an extra-light model for text segmentation, which can then be used for most complex tasks without being restrictive for deployment.

We are conscious that our models are not as competitive as the best deep learning ones, especially for NER. These latter use a large amount of data to construct word embeddings \cite{glove} \cite{bojanowski2017enriching} \cite{akbik2019pooled}, which become the input of models, like BiLSTM-CRF, for example, to achieve state-of-the-art results \cite{akbik2019flair}. On the one hand, using these embeddings is not possible with PMC with the Viterbi or the Forward-Backward algorithms, as this model cannot use observations with arbitrary features. On the other hand, these neural models are heavy and impossible to deploy without a significant configuration.

The training time, the execution time, and the carbon footprint of PMC are significantly lower, which is our project's main objective. As a perspective, PMC has been extended with the Triplet Markov Chain (TMC) model \cite{pieczynski2002chaines} \cite{chen2020modeling} \cite{li2019adaptive}. We can apply this extension to observe the possible improvements, especially for NER, while keeping Markov models' relevant properties.

\subsection*{About the Corresponding Author}

Elie Azeraf holds a Ph.D. in Statistics from Institut Polytechnique de Paris, as well as three Master of Science degrees in Financial Engineering, Applied Mathematics, and Data Science. His research interests include finance, time series analysis, and statistical modeling. He is the founder of Azeraf Financial Consulting \href{https://www.azeraf-financial-consulting.com/}{https://www.azeraf-financial-consulting.com/}, a firm dedicated to the design and implementation of portfolio construction strategies tailored to the specific objectives and risk profiles of individual and professional investors.

\bibliographystyle{unsrt}  
\bibliography{references}

\end{document}